# Network-wide link travel time and station waiting time estimation using automatic fare collection data: A computational graph approach

Jinlei Zhang, Feng Chen*, Lixing Yang, Wei Ma*, Guangyin Jin, and Ziyou Gao

*Abstract*—Urban rail transit (URT) system plays a dominating role in many megacities like Beijing and Hong Kong. Due to its important role and complex nature, it is always in great need for public agencies to better understand the performance of the URT system. This paper focuses on an essential and hard problem to estimate the network-wide link travel time and station waiting time using the automatic fare collection (AFC) data in the URT system, which is beneficial to better understand the system-wide real-time operation state. The emerging data-driven techniques, such as computational graph (CG) models in the machine learning field, provide a new solution for solving this problem. In this study, we first formulate a data-driven estimation optimization framework to estimate the link travel time and station waiting time. Then, we cast the estimation optimization model into a CG framework to solve the optimization problem and obtain the estimation results. The methodology is verified on a synthetic URT network and applied to a real-world URT network using the synthetic and real-world AFC data, respectively. Results show the robustness and effectiveness of the CG-based framework. To the best of our knowledge, this is the first time that the CG is applied to the URT. This study can provide critical insights to better understand the operational state in URT.

## I. INTRODUCTION

With the increasing urbanization and city agglomerations, urban rail transit (URT) plays an important role in urban transportation systems. To better operate and manage the complex URT system, understanding passengers' spatiotemporal congestion distributions in transit systems is in great need [1, 2]. To obtain the spatiotemporal congestion pattern, we need to estimate the link travel time, which refers to the in-vehicle time from one station to another adjacent station, as well as station waiting time, which refers to the waiting time at the origin stations and transfer stations. Both link travel time and station waiting time could help the public agencies track the spatiotemporal congestion patterns and identify the bottlenecks of the URT system. System performance measures of the URT system can be obtained from the link travel time and station waiting time, and hence they are the fundamental inputs for many URT applications[3-5]. Moreover, after obtaining the link travel time and station waiting time information, we can compute travel time of a specific path, which is the summation of several link travel times and station waiting times. With accurate path travel time, passengers can arrange their schedules [6], operators can conduct measures to avoid traffic congestions according to this information, and advertisers can post advertisements in subway stations according to the station waiting time [7].

Existing studies estimate the link travel time and station waiting time in the URT systems under simplified assumptions. For example, the most straightforward way to estimate the link travel time is to directly extract from the train timetable. However, the timetable is strictly confidential and can only be used for train scheduling by the operators in many cities. Even the timetable is available, it cannot be guaranteed that the trains could be punctual. For example, if passengers' clothes were stuck by a train gate, the train can be delayed for several seconds [8]. As for the station waiting time, existing studies generally treat half of the headway as the average station waiting time [9], which cannot reflect the actual traffic conditions. Overall, in many real-world URT systems, the dynamic and true station delay time are not available.

With the increasing popularity of the automatic fare collection (AFC) system in URT system, AFC data becomes accessible, and it contains the origin, destination, entry time, and exit time of each trip record. There is a great potential to extract useful travel time information from the AFC data. Therefore, it is of great interest and importance to develop models to estimate the accurate link travel time and station waiting time using AFC data.

Tremendous studies are using the AFC data to conduct different types of studies. Many researchers used AFC data to analyze the route choice model in URT systems [10-14]. The travel utility in behavior models can be estimated from AFC data using conventional statistical methods. The traffic assignment model was also thoroughly explored in URT systems with AFC data [17, 18]. Some studies dived into the

Manuscript received August 4, 2021; date of current version August 4, 2021. This work was supported by the Fundamental Research Funds for the Central Universities (No. 2021RC270), the National Natural Science Foundation of China (Nos. 71621001, 71825004） and a grant funded by the Hong Kong Polytechnic University (No. P0033933).

Jinlei Zhang, Lixing Yang, and Ziyou Gao are with the State Key Laboratory of Rail Traffic Control and Safety, Beijing Jiaotong University, No.3 Shangyuancun, Haidian District, Beijing 100044, China. (e-mail: zhangjinlei@bjtu.edu.cn; lxyang@bjtu.edu.cn; zygao@bjtu.edu.cn). *(Corresponding author: Feng Chen and Wei Ma.)*

Feng Chen is with the School of Civil Engineering, Beijing Jiaotong University, No.3 Shangyuancun, Haidian District, Beijing 100044, China. (e-mail: fengchen@bjtu.edu.cn).

Wei Ma is with the Department of Civil and Environmental Engineering, the Hong Kong Polytechnic University, Kowloon, Hong Kong SAR, China. (e-mail: wei.w.ma@polyu.edu.hk).

Guangyin Jin is with the College of System Engineering, National University of Defense Technology, Changsha 410005, China (e-mail: jinguangyin18@nudt.edu.cn).

Table 1 Comparison between similar studies

| Studies | Information estimated | Cases | Data | Supplementary means | Supplement data | Methods |
|---|---|---|---|---|---|---|
| Li et al. [9] | Walking and waiting time at origin stations | Four subway stations | AFC data | None | Train timetable | Clustering, Maximum Likelihood Method, Traffic assignment |
| Lee et al. [7] | Walking time, riding time (in-vehicle time), and waiting time | Two subway stations | AFC data | None | None | Clustering, Statistic method, Regressive method |
| Zhang et al [8] | Walking time, waiting time, in-vehicle time, and transfer time | Subway network | AFC data | Large-scale onsite investigation to obtain dwelling time | None | Clustering, Statistic method. |
| Wu et al. [15] | Walking time, in-vehicle time, and transfer time. | Subway network | AFC data | None | Train timetable | Clustering, Statistic method, normal distribution method. |
| Oh et al. [16] | Dwell time | One subway line (Totally twelve subway stations) | AFC data | None | Real-time metro operation data | Support vector regression |
| *This study* | *link travel time and station waiting time** | *Subway network* | *AFC data* | *None* | *None* | *Computational graph model* |

*Note that the link travel time includes the in-vehicle time and transfer walking time, and the station walking time includes the origin station walking time and transfer station walking time.

commuting pattern or travel pattern analysis [19-21], and short-term passenger flow predictions [22-25] using the AFC data. The passengers' trajectories were also estimated via AFC data [26, 27].

However, there are relatively few studies to estimate the link travel time and station waiting time using AFC data. Li et al. [9] estimated the waiting and walking time in subway stations using AFC data. By assigning passengers to different trains, they estimated the walking time distribution, which was used to estimate the waiting and walking time. In their study, they estimate for a small portion of the stations instead of the entire subway network. Lee et al. [7] decomposed the travel time into walking time, riding time (in-vehicle time), and waiting time in subway networks, which is similar to our study. Besides, a statistic- and regression-based method is developed to conduct the estimation on a few stations. In our study, we further divide the riding time into link travel times between stations. This kind of partition method is more meaningful because we can get more fine-grained link travel time information. A computational graph (CG)-based method is also conducted to conduct the network-wide estimation. Zhang et al [8] segmented the travel time into walking time, waiting time, in-vehicle time, and transfer time. They conducted a field survey to obtain the train dwell time, which is supplementary means for the estimation process. Wu et al. [15] divided the total travel time into walking time, in-vehicle time, and transfer time. The timetable data is available in their study. The similarity of these two studies was that they firstly estimated the walking time using non-transfer AFC records, and then estimated the transfer time using transfer AFC data. Oh et al. [16] incorporated the AFC data and train operation data to estimate the dwell time via the support vector regression-based model. However, the model is specifically for Seoul, South Korea because of the data specificity. All in all, existing travel time estimation-related studies only take several OD pairs or a subway line as examples. Train timetable or other real-time operation data are generally supplementary data used in the estimation process, and conventional models are adopted, which is difficult to scale up for the network-wise estimation.

To highlight the uniqueness of this paper, differences between existing studies and this study are summarized in Table 1.

The emerging data-driven computational graph (CG) model in the machine learning area provides a new solution scheme for the estimation of link travel time and station waiting time on a network scale. The CG is a type of computational framework, including nodes and edges. The nodes represent variables or operations on variables. The variables can be scalars, vectors, matrices, or tensors, etc. The operations on variables can be a recurrent neural network, conventional neural network, or fully connected network, etc. The edges represent the data dependency between variables, that is, a flow relationship between variables. Therefore, the CG can also be called a data flow graph. Fig. 1 shows a CG with two fully connected layers in PyTorch. It transforms the input tensor of (64, 1000) dimension into a tensor of (64, 100) dimension through the first fully connected layer, and then into a tensor of (64, 10) dimension through the second fully connected layer, which is the final output. The backward propagation algorithm in machine learning is based on the chain rule. In strict accordance with the data flow relationship in the CG, the error is backpropagated from the output node to the input node, so as to optimize parameters on the computational graph.

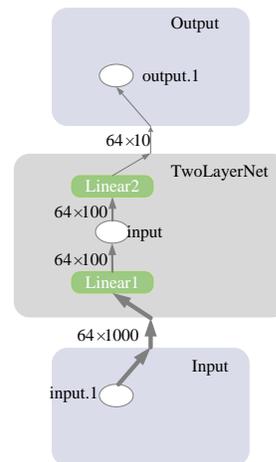

Fig. 1 Diagram of the computational graph

In recent years, CG-based models have been widely used in various transportation applications [28]. Some studies began to cast the conventional four-step traffic demand model and OD estimation problem into the CG framework. Leveraging the backward propagation algorithm in TensorFlow or PyTorch, the CG can solve the specific parameters in the model. For example, Wu et al. [29] cast household travel surveys, mobile phone sample data, sensor data, floating car data and vehicle location/identification data into a hierarchical CG framework. Through the forward propagation of multiple data sources and the backward propagation of the loss errors of travel time, many demand variables in the three of the four-step traffic demand model (namely trip generation, spatial distribution estimation, and path flow-based traffic assignment) were estimated. This framework has favorable interpretation and was beneficial to better understand the traffic behaviors. Similar to this study, Sun et al. [30] further analyzed the marginal effects of all kinds of traffic policies, e.g., adding the toll on links, adding flows to OD pairs, removing flows from OD pairs. The marginal effects and potential traffic migration caused by various traffic strategies were quantified by casting the various traffic policies individually or cooperatively into the above hierarchical CG framework. Ma et al [31] solved the dynamic OD estimation problem under a large-scale road network using the CG framework. Through the forward propagation of initialized OD demand and the backward propagation of the loss errors of link flows or link speeds, many parameters used in the dynamic traffic assignment models and route choice models, such as OD demand, route choice probability, dynamic assignment ratio, and link/path travel time, were estimated. In summary, the data-driven CG framework provides a powerful and promising solution for the scenario of traffic problems. However, existing related models are all about the road network. There is no study using the CG framework in URT.

To summarize, there are several challenges in the estimation of link travel time and station waiting time in URT. First, the timetable data and real-time operation data may not be available. Therefore, how to estimate this information using only the AFC data is challenging. Second, with the increasing scale of URT systems, it is impractical to conduct an onsite investigation (or field survey) to obtain some supplementary information to assist the estimation process. Thirdly, existing models generally consist multiple sub-modules in the estimation process, which makes the estimation complicated. How to incorporate the complicated estimation process in an end-to-end framework also deserves to be explored.

To this end, we apply the CG framework to the link travel time and station waiting time estimation problem in the URT system. We construct one CG framework only using the AFC data to estimate the link travel time and station waiting time under a network level. Firstly, we formulate the data-driven estimation problem for link travel time and station waiting time. Then, we vectorize the variables in the estimation problem and cast them in the CG framework. Finally, the methodology is verified using a synthetic URT network as well as the real-world URT network in Beijing, China. The contributions of this study are summarized as follows.

(1) Making use of the AFC data, we formulate a data-driven framework to estimate the link travel time and station waiting time for all the stations and links on a large URT network. No other supplementary means or data are needed in the framework.
(2) It is the first time that the computational graph model is applied to URT systems. The proposed framework is end-to-end in the sense that the AFC data is input into the CG based-framework and the link travel time and station waiting time can be directly estimated.
(3) The proposed framework is strictly verified on a synthetic URT network and is applied to the real-world URT network in Beijing, China.

The remainder of this paper is organized as follows. We define the problem and notations in Section II. In Section III, we present the estimation problem and CG framework formulation. The case study is conducted in Section IV, including the synthetic case and the real-world case. Finally, we draw findings and conclusions in Section V.

## II. PROBLEM DEFINITION

### A. Travel time decomposition

Each trip of a passenger from tap-in to tap-out in the URT system can be represented into several phases as shown in Fig. 2 [15]. In the origin station, the passenger will tap in and walk to the platform to wait for the next train. If there are queues, he/she cannot board the closest train and have to wait for another train. Therefore, the waiting time will depend on the queue length. After departing from the origin station, the passenger will take line 1 to go to the transfer station, if any. In the transfer station, he/she will walk to another platform and wait for the next train. Similar to the origin station, there is a waiting time. After departing from the transfer station, the passenger will take line 2 to go to the destination station and walk to tap out.

The complete trip can be encapsulated into the subway network graph $G = (V, E)$, where $V$ is the set of all nodes, including the gate machine node and platform node. $E$ is the set of all links, including the entry link, exit link, vehicle link, and transfer link. All links are associated with the direction of uplink and downlink. The links from different directions have different attributes. In summary, a complete trip includes four types of links and two types of nodes. The meanings of the notations are listed as follows.

(1) Entry link: The links from the gate machine to the platform.
(2) Vehicle link: The links that passengers are in the vehicle.
(3) Transfer link: The links that passengers transfer from one subway line to another subway line at transfer stations.
(4) Exit link: The links from the platform to the gate machine.
(5) Gate machine node: The node at the gate machine.
(6) Platform node: The node at the platform.

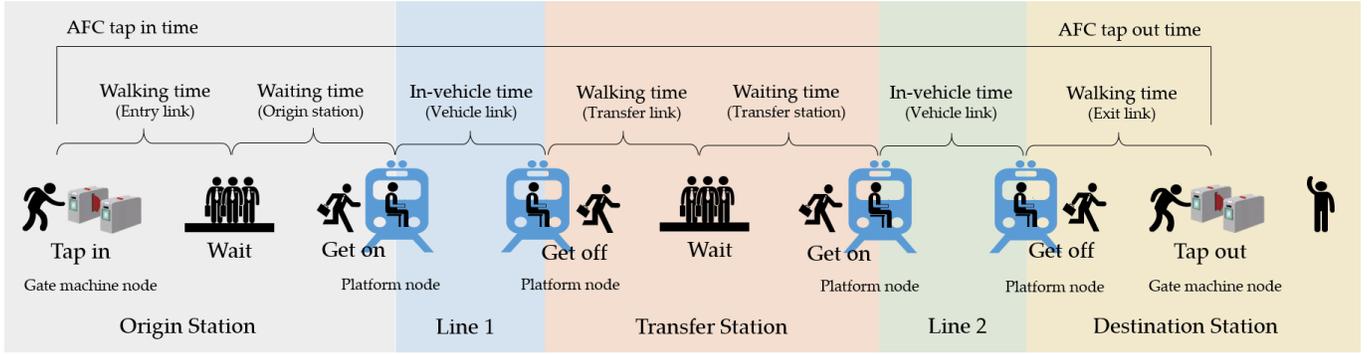

Fig. 2 Diagram of passengers' trip behavior

(7) Uplink and downlink: If the direction of link a→b is the uplink, the direction of link b→a is the downlink.

The AFC data can record the card number, entry station, entry time, exit station, exit time. Therefore, the true travel time of a single trip is available.

*B. Network representation*

In order to further illustrate the problem, a synthetic URT network with 5 subway stations and two subway lines is constructed as shown in Fig. 3. The green line is line 1, including 3 stations (stations 1, 2, 3). The blue line is line 2, including 3 stations (stations 4, 2, 5). In the diagram, the blue nodes are general stations and the green node is a transfer station.

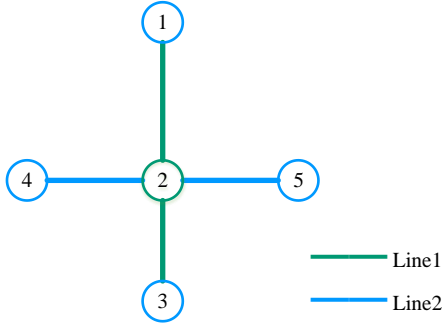

Fig. 3 Diagram of the synthetic URT network

The station can be divided into the general station and transfer station. All stations are numbered according to the line number and station adjacency relationship. During the network modeling process, we divide the general station into four nodes, namely one gate machine node, one platform node, and two train gate nodes, as shown in Fig. 4. The number of the gate machine is the original station number with the letter G, such as $V_1^G, V_2^G, V_3^G$. The number of the platform node is the original station number with the letter P, such as $V_1^P, V_2^P, V_3^P$. The number of the train gate nodes are the original station number with the letter $T_1$ and $T_2$, indicating different directions, such as $V_1^{T_1}, V_1^{T_2}$. We divide the transfer station into seven nodes, denoting the gate machine node $V_2^G$, the original platform node $V_2^{PA}$, the virtual platform node $V_2^{PB}$, and four train gate nodes $V_2^{TA_1}, V_2^{TA_2}, V_2^{TB_1}, V_2^{TB_2}$, as shown in Fig. 4.

The links in the synthetic are divided into entry link, exit link, vehicle link, and transfer link, as shown in Fig. 4. Every link includes the attributes of the origin, destination, direction, line number, travel time, and link name.

*C. Problem statement*

In this study, we will estimate the link travel time and the station waiting time, as shown in Fig. 4, only using the AFC data. The link travel time includes walking time on the entry link, transfer link, and exit link, as well as the in-vehicle time on the vehicle link. The station waiting time includes waiting time at the platform node of the origin station or transfer station. The problem is cast into the CG framework, with inputs being the AFC data and outputs being the time of all information mentioned above.

*D. An intuitive example*

Take path 1 (station 4 to station 2) and path 2 (station 4 to station 5) as shown in Fig. 5 as an example, we will estimate the waiting time $t_v^h$ at the node $V_4^P$ of the station 4, as well as the link travel time of the entry link, vehicle link, and exit link only using the AFC data. For path 1 and path 2, passengers only wait at station 4. Therefore, the station waiting time only exists at station 4. Passengers tap in at node $V_4^G$ and wait at node $V_4^P$ to board. Some passengers alight at node $V_2^{TB_2}$ and tap out at node $V_2^G$, and others alight at node $V_5^{T_2}$ and tap out at node $V_5^G$. Hence, the travel time of five links needs to be estimated. In brief, we will estimate the link travel time (five links) and station waiting time (one station) for path 1 and path 2.

*E. Notations*

All notations are summarized in Table 2.

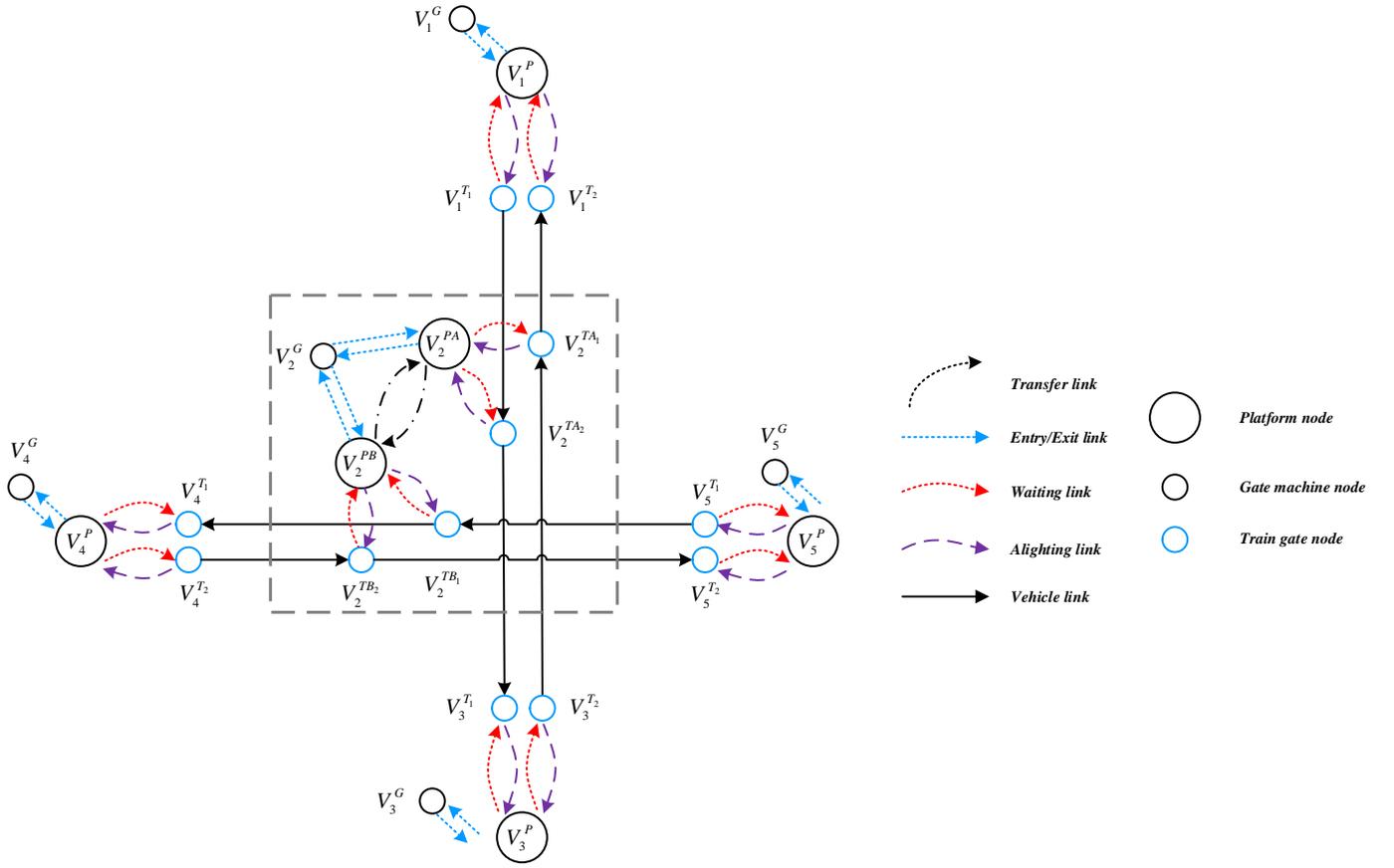

Fig. 4 Diagram of network representation

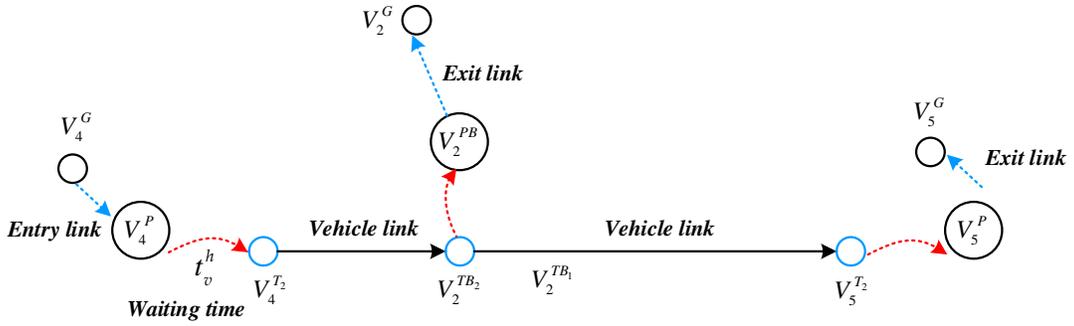

Fig. 5 An example of the problem statement

Table 2 Notations in the framework

| Notations | Definitions |
|---|---|
| $G$ | The URT network graph |
| $V$ | The set of all nodes in the URT network |
| $E$ | The set of all links in the URT network |
| $a \in E$ | The link in the URT network |
| $v \in V$ | The node in the URT network |
| $r$ | The origin station |
| $s$ | The destination station |
| $K_q$ | The set of all OD pairs |
| $k$ | The $k_{th}$ shortest path |
| $K_{rs}$ | The set of $k$ shortest paths from station $r$ to $s$ |
| $h \in H=(h_1, h_2, h_3\ldots)$ | The time interval that passengers enter the URT network |
| $t_a$ | The link travel time that passengers pass the link $a$ |
| $t_v^h$ | The passengers' waiting time at station $v$ during the time interval $h$ |
| $\alpha_{rs}^{ka}$ | The coefficient of $t_a$ |
| $\beta_{rs}^{kvh}$ and $\gamma_{rs}^{kvh}$ | The coefficient of $t_v^h$ |
| $c_{rs}^{kh}$ | The utility function |
| $\psi_{rs}^{kh}$ | The generalized route choice model |
| $\theta$ | The parameter of the logit model |
| $p$ | The probability that passengers choose the $k$ path |
| $\tilde{c}$ | The vector of the true travel time |
| $B$ | The vector of the link |
| $M^h$ | The vector of the node |
| $A$ | The combined vector of the link and node |
| $t_a$ | The vector of the link travel time |
| $t_v^h$ | The vector of the link travel time and station waiting time |
| $t$ | The combined vector of the link travel time and the station waiting time |
| $P$ | The vector of the route choice probability |

## III. METHODOLOGY

In this section, we present the estimation framework in large-scale URT networks. The framework comprises two parts:

(1) Define the estimation framework on the basis of the passengers' route choice model; (2) Variable vectorization and development of a CG framework to solve this problem.

*A. Passengers' route choice model*

In Section 2.1, we have described the passengers' behavior during a single trip. Before choosing this route, passengers' route choice behaviors are affected by many factors including the walking time on the entry link and exit link, in-vehicle time on the vehicle link, transfer walking time on the transfer link, waiting time at the platform nodes of the origin station or transfer station, transfer times, and ticket fare, etc. In this study, we choose the link travel time and station waiting time as the generalized utility function, as shown in Eqs. (1).

$$c_{rs}^{kh} = \sum_a \alpha_{rs}^{ka} t_a + \sum_v \beta_{rs}^{kvh} t_v^h + \sum_v \gamma_{rs}^{kvh} t_v^h \quad (1)$$

where $c_{rs}^{kh}$ is the generalized utility, $rs \in K_q$, $K_q$ is the set of OD pairs, $k \in K_{rs}$, $K_{rs}$ is the path set from the origin station $r$ to destination station $s$, $h \in H = (h_1, h_2, h_3...)$ is the time interval that passengers enter the URT network, $a \in E$ is the link edge set in the URT network graph $G$, $v \in V$ is the station node set in the URT network graph $G$, $t_a$ is the travel time that passengers pass the link $a$, $t_v^h$ is the passengers' waiting time at station $v$ during the time interval $h$, $\alpha$、$\beta$ and $\gamma$ are the coefficient of $t_a$ and $t_v^h$. They are defined as follows.

The $\alpha_{rs}^{ka}$ equals 1 if the path $k$ of OD pair $rs$ passes the link $a$, otherwise equals 0.

The $\beta_{rs}^{kvh}$ equals *1* if the path $k$ of OD pair $rs$ boarding at the origin or transfer station $v$ in the uplink during time interval $h$, otherwise equals 0.

The $\gamma_{rs}^{kvh}$ equals 1 if the path $k$ of OD pair $rs$ boarding at the origin or transfer station $v$ in the downlink during time interval $h$, otherwise equals 0.

Several meanings should be noted in the generalized utility function.
(1) The $h$ is the time interval that passengers enter the URT network. Hence, the $c_{rs}^{kh}$ is associated with the time interval $h$. For the same route, there is different $c_{rs}^{kh}$ in different time intervals.
(2) The $t_v^h$ is also associated with the time interval $h$. For the same route, there is a different $t_v^h$ in different time intervals.
(3) The $t_v^h$ is associated with the direction of the train. For the same station in the same time interval, there is a different $t_v^h$. Therefore, the coefficients of $t_v^h$ for uplink and downlink are $\beta_{rs}^{kvh}$ and $\gamma_{rs}^{kvh}$, respectively.
(4) We assume that the walking time and in-vehicle time is independent of the time interval. Hence, $\alpha_{rs}^{ka}$ and $t_a$ is not associated with the time interval $h$.
(5) We assume that the stochastic utility is subject to the independent identical distribution and Gumbel distribution. Therefore, the route choice model can be defined as Eq. (2).

$$p_{rs}^{kh} = \Psi_{rs}^{kh}(c, t) \quad (2)$$

$$c = \{c_{rs}^{kh} | h \in H, rs \in K_q, k \in K_{rs}\}$$

$$t = \{t_a, t_v^h | h \in H, a \in E, v \in V\}$$

where $p_{rs}^{kh}$ is the probability that passengers choose the $k$ path between OD pair $rs$ within the time interval $h$. $\Psi_{rs}^{kh}$ is a generalized route choice model that is used to compute the probability. $c_{rs}^{kh}$ is the utility function as shown in Eq. (1), including the link travel time $t_a$ and station waiting time $t_v^h$.

*B. Estimation framework: A data-driven optimization model*

In this section, we will formulate the estimation framework leveraging the AFC data. Through minimizing the difference between the estimated OD travel time and the true OD travel time, we will estimate the link travel time (including the walking time when entering or exiting the station and in-vehicle time) and station waiting time (including the waiting time at the origin station and the transfer station). Therefore, we will first formulate the true OD travel time and the estimated OD travel time.

*C. The true OD travel time*

For one piece of the AFC data, let the entry station, entry time, exit station, exit time be the $r$, $\tilde{t}_{ir}$, $s$, $\tilde{t}_{is}$, respectively. Therefore, the true travel time can be formulated as Eq. (3).

$$\tilde{c}_{irs}^h = \tilde{t}_{is} - \tilde{t}_{ir} \quad (3)$$

where $\tilde{c}_{irs}^h$ is the true travel time of passenger $i$ for the OD pair $rs$. Because the AFC data cannot record that which path of the OD pair $rs$ the passengers chose, the $\tilde{c}_{irs}^h$ is not associated with the path $k$ of the OD pair $rs$.

There are many records for the same OD pair in the same time interval. Among these records, the travel time is generally different because of the different entry time and different routes. To reduce the computational cost, we choose the true average OD travel time as the true OD travel time, as Eq. (4).

$$\tilde{c}_{rs}^h = \frac{1}{N_{rs}^h} \sum_{i=1}^{N_{rs}^h} \tilde{c}_{irs}^h = \frac{1}{N_{rs}^h} \sum_{i=1}^{N_{rs}^h} (\tilde{t}_{is} - \tilde{t}_{ir}) \quad (4)$$

where $N_{rs}^h$ is the record number of the OD pair $rs$ in the time interval $h$, $\tilde{c}_{irs}^h$ is the true travel time for the specific OD pair $rs$ in the time interval $h$.

*D. The estimated OD travel time*

Given the route choice probability $p_{rs}^{kh}$ and travel utility $c_{rs}^{kh}$, the estimated average OD travel time of the OD pair $rs$ in the time interval $h$ is shown as Eq.(5). When estimating the OD travel time, we should consider the probability of different paths of the OD pair $rs$. Therefore, the $c_{rs}^{kh}$ is associated with the path $k$ of the OD pair $rs$.

$$c_{rs}^h = \sum_k p_{rs}^{kh} c_{rs}^{kh} \quad (5)$$

*E. The optimization model*

The objective is to minimize the difference between the estimated OD travel time and the true OD travel time, as shown in Eq. (6)

Table 3 Variable vectorization

| Variable | Scalar | Vector | Dimension | Type |
|---|---|---|---|---|
| True travel time | $\tilde{c}_{r_i s_i}^h$ | $\tilde{c}$ | $R^{(N_{rs} \times H) \times 1}$ | Dense |
| Links | $\alpha_{rs}^{ka}$ | $B$ | $R^{(N_{rs} \times k) \times N_a}$ | Sparse |
| Nodes | $\beta_{rs}^{kvh}, \gamma_{rs}^{kvh}$ | $M^h$ | $R^{(N_{rs} \times k) \times N_v}$ | Sparse |
| Links and nodes | $\alpha_{rs}^{ka_1}, \beta_{rs}^{kvh}, \gamma_{rs}^{kvh}$ | $A$ | $R^{(N_{rs} \times k \times H) \times (N_a + N_v \times H)}$ | Sparse |
| Link travel time | $t_a$ | $t_a$ | $R^{N_a \times 1}$ | Dense |
| Station waiting time | $t_v^h$ | $t_v^h$ | $R^{N_v \times 1}$ | Dense |
| Variable to be estimated | $t_a 、 t_v^h$ | $t$ | $R^{(N_a + N_v \times H) \times 1}$ | Dense |
| Route choice probability | $p_{rs}^{kh}$ | $P$ | $R^{(N_{rs} \times H) \times (N_{rs} \times k \times H)}$ | Sparse |

$$\min \sum_{rsh} |\tilde{c}_{rs}^h - c_{rs}^h|_p^2 \quad (6)$$

Considering the Eqs. (1)-(6), the optimization model used in this study can be formulated as Eq. (7). The objective function is to minimize the difference between the estimated OD travel time and the true OD travel time. The decision variables are the link travel time $t_a$ and the waiting time $t_v^h$.

$$\min_{\{t_a\}_a, \{t_v^h\}_{v,h}} \sum_{rsh} \left| \tilde{c}_{rs}^h - \sum_k p_{rs}^{kh} c_{rs}^{kh} \right|_p^2 + |\tilde{t}_a - t_a|_p^2 + |\tilde{t}_v^h - t_v^h|_p^2$$

s.t. $\quad t_a \geq 0, a \in E$ (7)
$\quad t_v^h \geq 0, v \in V$
Eqs. (1)-(2)

### F. Variable vectorization

Before casting the optimization model into a CG framework to obtain the optimal solution, we will vectorize the variables used in Eq. (7), including the true travel time $\tilde{c}_{rs}^h$, the links $a$, the nodes $v$, the link travel time $t_a$, the station waiting time $t_v^h$, and the route choice probability $p$. After vectorization, the supervised tasks under the CG framework are constructed.

The summary of all variable vectorizations is shown in Table 3. The detailed vectorizing process is shown in the supplementary file.

Based on Table 3, the optimization model in Eq. (7) can be reformulated as Eq. (8).

$$\min_{\{t\}, \theta} \|\tilde{c} - PAt\|_2^2$$

s.t. $\quad t \geq 0$ (8)
$\quad P = logit(A; t; \theta)$

The main decision variable in the above optimization model is $t$, including the link travel time (walking time when entering or exiting the station and in-vehicle time) and station waiting time (waiting time at origin station or transfer station). When the matrix $A$ is with full rank, there is the only solution for the optimization problem.

### G. A computational graph for the optimization model

In order to solve the optimization model, we cast it into the CG framework. Fig. 6 shows the CG framework of this problem. The framework consists of two parts, the forward propagation part and the backward propagation part.

The forward propagation part is shown as the solid line in Fig. 6. During forward propagation, we assume that the station waiting time and link travel time are fixed. By calculating the path travel time and route choice probability, the estimated OD travel time is obtained, and the estimated OD travel time is compared with the true one to obtain the estimated error. In the backward propagation part, which is shown as the dashed line in Fig. 6, the error is backward propagated according to the original path to adjust the station waiting time and link travel time. In this way, repeated iteration calculation is carried out until the iteration stops, and the final estimated station waiting time and link travel time are obtained.

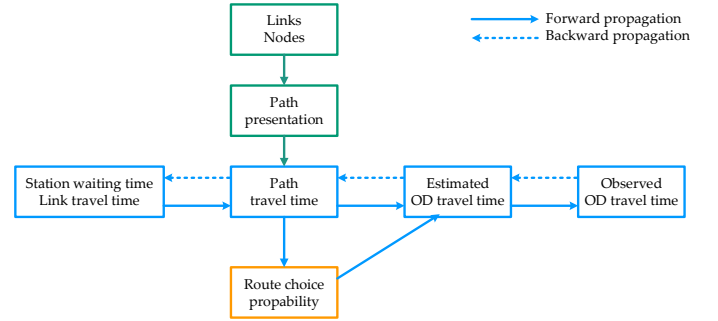

Fig. 6 The computational graph framework: a diagram of the forward-backward algorithm

Fig. 7 shows the corresponding variable propagation process. The solid line represents the forward propagation, and the dashed line represents the backward propagation. Through the forward propagation of variables and the backward propagation of errors, the final $t$ composed of station waiting time and link travel time is obtained.

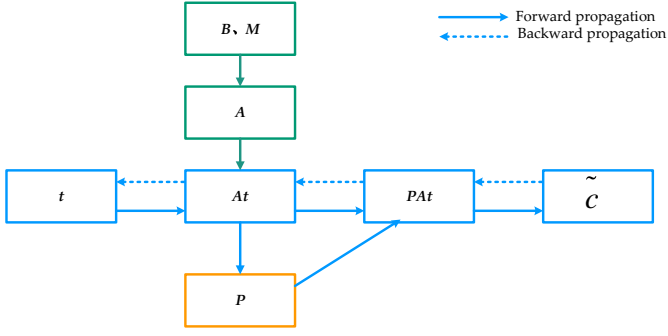

Fig. 7 The diagram of the forward-backward algorithm among all variables

***Forward propagation***: The input is the randomly initialized $t$ composed of the station waiting time $t_v^h$ and link travel time $t_a$. Together with the path representation $A$, the path travel time $c_{path}$ is obtained. According to the path travel time, the route choice probability $P$ is obtained. The path travel time $c_{path}$ multiply the route choice probability $P$ to get the estimated OD travel time.

The objective of the forward propagation is to obtain the estimated OD travel time. Let the objective function of Eq. (8) be the loss function in the machine learning field, the Eq. (8) can be decomposed into Eq.(9).

$$L = \|\tilde{c} - c\|_p^2$$

$$\begin{aligned} c &= P c_{path} \\ P &= \text{softmax}(-\theta c_{path}) \\ c_{path} &= At \end{aligned} \quad (9)$$

where $\tilde{c}$ is the true average OD travel time extracted from the AFC data, $c$ is the estimated average OD travel time, $P$ is the route choice model obtained using the logit model, $c_{path}$ is the path travel time.

***Backward propagation***: During this process, we take the route choice probability $P$ as a known variable. Then, Eq. (8) can be simplified into Eq. (10).

$$\min_{\{t\},\theta} \ \|\tilde{c} - PAt\|_2^2 \quad (10)$$
$$s.t. \quad t \geq 0$$

Further, the gradient of the objective function can be obtained as shown in Eq. (11).

$$\begin{aligned} \frac{\partial L}{\partial c} &= 2(\tilde{c} - PAt) \\ \frac{\partial L}{\partial c_{path}} &= P^T \frac{\partial L}{\partial c} \\ \frac{\partial L}{\partial t} &= A^T \frac{\partial L}{\partial c_{path}} \end{aligned} \quad (11)$$

By consolidating Eq. (11), the gradient of the objective function $L$ against $t$ can be obtained, as shown in Eq.(12). Finally, the problem of the Eq. (8) can be solved leveraging the backpropagation algorithm.

$$\frac{\partial L}{\partial t} = A^T P^T 2(\tilde{c} - PAt) \quad (12)$$

IV. CASE STUDY

In this section, we first present the synthetic and real-world data. The model configuration is then described. The result analyses are finally showed.

*A. Data Description*

*1) Synthetic data*

In order to verify the proposed framework, a synthetic URT network is constructed. The network modeling method is the same as that in Section 2. Fig. 8 shows the synthetic subway network with 3 subway lines, including one ring line and 2 general lines, and 14 subway stations, including 8 general stations and 6 transfer stations. The green line is line 1, including 7 stations (stations 1, 2, 3, 4, 5, 6, 7). The yellow line is the ring line 2, including 5 stations (station 8, 9, 104, 10, 102). Station 104 and station 102 are the same station with station 4 and station 2, respectively. The blue line is the line 3, including 8 stations (station 11, 109, 103, 110, 12, 13, 106, 14). Station 109, 103, 110, and 106 are the same station with the station 9, 3, 10, 6. In the diagram, the blue nodes are the general stations and the green nodes are the transfer stations.

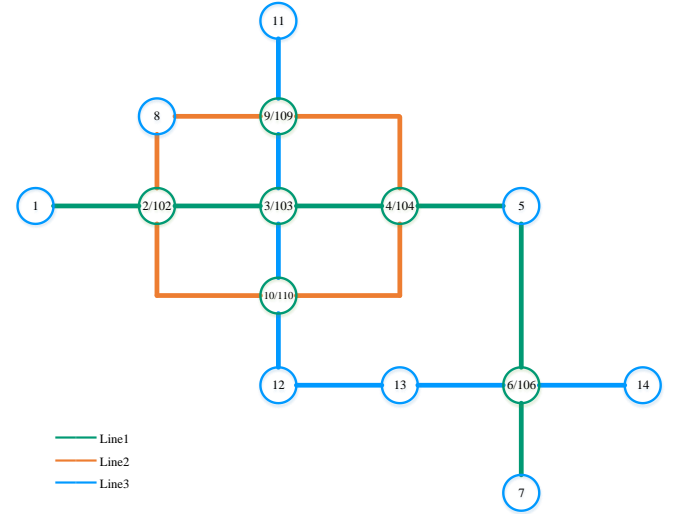

Fig. 8 Diagram of the synthetic URT network

In the synthetic network, the link travel time and station waiting time are artificially generated. According to this information, the synthetic AFC data are artificially constructed. Therefore, all the true traffic state information in this synthetic network is available.

The synthetic AFC data is shown in Table 4. The travel time including theoretical true travel time and travel time with errors. The theoretically true travel time does not include any noises. The travel time with white noises is obtained by applying a random noise subject to normal distribution to the theoretically true travel time in order to simulate the real travel environment.

When constructing AFC data, note that: (1) To verify the proposed methodology, we construct enough AFC data for all OD pairs. However, when conducting sensitivity analysis, we will delete some data records of some OD pairs to simulate the actual AFC data. (2) The number of data records can be different for different OD pairs. (3) For the same OD pair, the

Table 4 Example of synthetic AFC data

| | Card ID | Entry station number | Exit station number | Entry time | Exit time | Theoretically true travel time | Travel time with white noise |
|---|---|---|---|---|---|---|---|
| 1 | 1000 | 5 | 10 | 55 | 65.356 | 10.356 | 10.785 |
| 2 | 1000 | 3 | 4 | 36 | 38.691 | 2.691 | 3.001 |
| 3 | 1000 | 1 | 14 | 48 | 91.603 | 43.603 | 44.891 |
| … | … | … | … | … | … | … | … |

Table 5 Example of synthetic AFC data for the same OD pair and different $k$ paths

| | Card ID | Entry station number | Exit station number | Entry time | Exit time | Theoretically true travel time | Travel time with white noise | Card ID |
|---|---|---|---|---|---|---|---|---|
| 1 | 1000 | 13 | 9 | 82 | 99.393 | 17.393 | 18.005 | Route 1 |
| 2 | 1000 | 13 | 9 | 24 | 46.497 | 22.497 | 23.001 | Route 2 |
| 3 | 1000 | 13 | 9 | 38 | 63.368 | 25.368 | 26.345 | Route 3 |
| 4 | 1000 | 14 | 3 | 55 | 72.676 | 17.676 | 18.596 | Route 1 |
| 5 | 1000 | 14 | 3 | 285 | 304.184 | 19.184 | 21.587 | Route 2 |
| 6 | 1000 | 14 | 3 | 164 | 194.54 | 30.540 | 32.998 | Route 3 |
| … | … | … | … | … | … | … | … | … |

Table 6 Example of processed AFC data

| | Card ID | Entry station number | Exit station number | Entry time | Exit time | Travel time |
|---|---|---|---|---|---|---|
| 1 | 74873*** | 19 | 37 | 849 | 875 | 26 |
| 2 | 19727*** | 44 | 43 | 452 | 466 | 14 |
| 3 | 42656*** | 29 | 43 | 356 | 464 | 108 |
| … | … | … | … | … | … | … |

Table 7 Example of the link distance information

| Origin station number | Destination station number | Direction | Line | Origin station name | Destination station name | Link distance |
|---|---|---|---|---|---|---|
| 1 | 2 | 0 | 1 | Pingguoyuan | Gucheng | 2606 |
| 2 | 3 | 0 | 1 | Gucheng | Bajiaoyouleyuan | 1921 |
| 4 | 3 | 1 | 1 | Babaoshan | Bajiaoyouleyuan | 1953 |
| 5 | 4 | 1 | 1 | Yuquanlu | Babaoshan | 1479 |
| 10001 | 1 | 2 | 10000 | Pingguoyuan | Pingguoyuan | 1000 |
| 10002 | 2 | 2 | 10000 | Gucheng | Gucheng | 1000 |

Note: For the direction, 0 means uplink, 1 means downlink, 2 means transfer, entry, and exit link. The line number for the transfer link is 100. The line number for the entry and exit link is 10000.

data records for different $k$ paths are generated according to the logit model. (4) For the same OD pair, the number of data records varies in different time intervals in order to simulate the demand changes in peak hours and off-peak hours. (5) To verify the proposed methodology, the travel time is the theoretically true travel time without white noises. However, when conducting sensitivity analysis, we will use the travel time with white noises to simulate the travel time obtained from the actual AFC data. The example of synthetic AFC data for the same OD pair and different $k$ paths is shown in Table 5.

*2) Real-world data*

The dataset from Beijing Subway in China on March 7, 2016 is used in the experiment, as shown in Table 6. The travel time is obtained by subtracting the entry time from the exit time, and the station number is the unique for each station in Beijing. The entry time and exit time are converted to minutes (There are 1440 minutes throughout the day). The URT network in Beijing is also available, which includes link distance information between adjacent stations and the transfer walking time. We set the average train speed of the Beijing subway is 30 km/h. We further translate the walking time into the driving distance of the train to search for the *k* shortest path. An example of the link distance information is shown in Table 7

### B. Model configuration

For the synthetic data, the time period is 5 hours from 7:00 am to 12:00 am, and each time interval is 30 minutes, a total of 10 time intervals. For the 182 OD pairs, there are 18.1 million AFC records. The optimizer used is AdaGrad. The learning rate is 0.1. The batch size is 8. The loss function is the mean square error (MSE). The loss variation is shown in Fig. 9. It can be seen that when 20 epochs are reached, the model has remained stable with a good convergence effect.

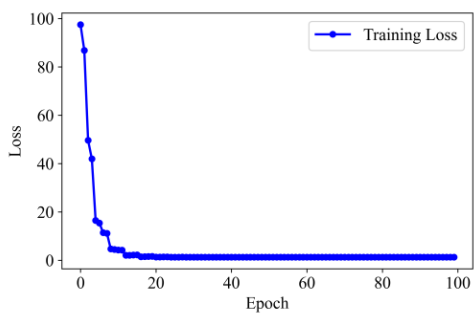

Fig. 9 The convergence curve of training loss

For the real-world data, the time period is also 5 hours from 7:00 am to 12:00 am, and each time interval is 30 minutes, a total of 10 time intervals. For the 767,176 OD pairs, there are 20.9 million AFC records. The optimizer used is AdaGrad. The learning rate is 0.1. The batch size is 1024*5. The loss function is the MSE. The loss variation is shown in Fig. 10. It can be seen that when 10 epochs are reached, the model has remained stable with a good convergence effect.

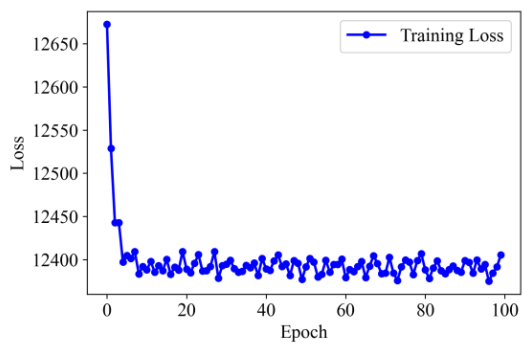

Fig. 10 The variation of training loss

### C. Result analysis for synthetic subway networks

In this section, we will verify the methodology using the synthetic subway network and the synthetic AFC data. For the synthetic network, all the true traffic states are available, including the true link travel time, the true station waiting time, and the true path travel time. The estimated *t* includes the estimated link travel time and station waiting time in different time intervals. Leveraging these estimated link travel time and station waiting time, we can reconstruct the estimated path travel time. Therefore, the estimated and the true link travel time, station waiting time, and path travel time are all available. The summary of this information is shown in Table 8.

Table 8 The information summary for the synthetic subway network

|  | Estimated values | True values |
|---|---|---|
| Link travel time | √ | √ |
| Station waiting time | √ | √ |
| Path travel time | √ | √ |

#### 1) Link travel time and station waiting time

The comparison between the true and the estimated link travel time and station waiting time (delay time) is shown in Fig. 11. The estimated values and the true values are also generally equal to each other. The $R^2$ reaches up to 0.999, which proves the effectiveness of the proposed methodology.

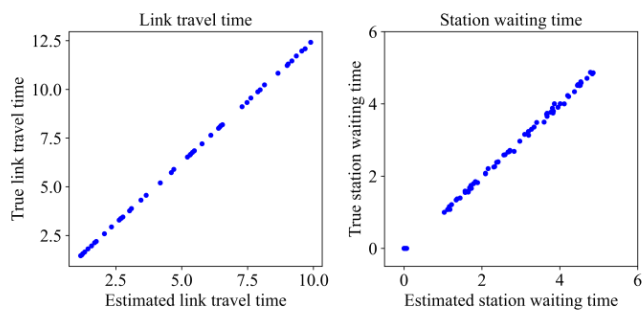

Fig. 11 The comparison between the estimated and the true time (unit: mins)

#### 2) Path travel time

The comparison between the true and the estimated path travel time is shown in Fig. 12. The estimated values and the true values are generally equal to each other. The $R^2$ reaches up to 0.999, which proves the effectiveness of the proposed methodology.

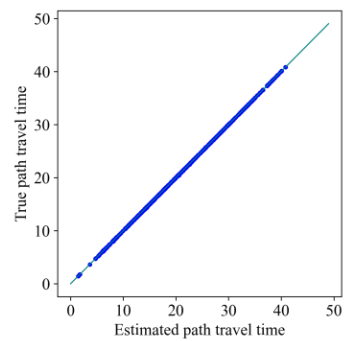

Fig. 12 The comparison between the estimated and the true OD path travel time (unit: mins)

#### 3) Sensitivity analysis

For the real-world AFC data, the path travel time always exists noises. Moreover, there is no AFC record for many OD pairs. To simulate the real-world AFC data, we conduct the sensitivity analysis in this section. At first, we add 10% and 20% white noise to the path travel time of all AFC records. On the basis of the 20% white noise, we randomly delete the records of 20% and 50% OD pairs. The results are presented as follows.

(1) Adding 10% white noises to the path travel time

In this section, we add 10% white noise to the path travel time. The estimated results are shown in Fig. 13 and Fig. 14. The error of the station waiting time becomes larger. The estimated link travel time and the path travel time are still aligned with their corresponding true values. Results show that the proposed methodology is effective in this case.

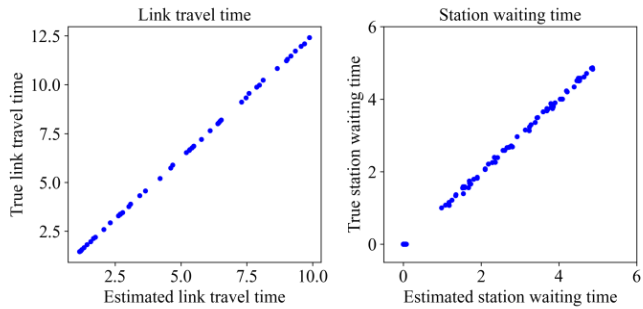

Fig. 13 The comparison between the estimated and the true time (unit: mins)

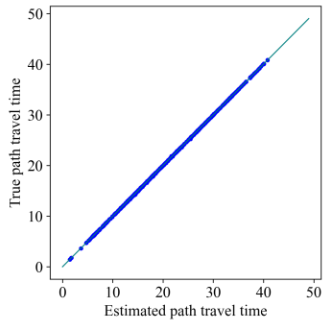

Fig. 14 The comparison between the estimated and the true OD path travel time (unit: mins)

(2) Adding 20% white noises to the path travel time

In this section, we add 20% white noise to the path travel time. The estimated results are shown in Fig. 15 and Fig. 16. Both of the errors of the link travel time and station waiting time become larger. The estimated link travel time still aligned with their corresponding true values. Results show that the proposed methodology is also effective in this case.

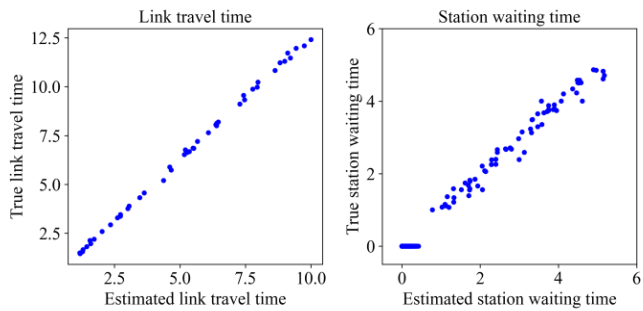

Fig. 15 The comparison between the estimated and the true time (unit: mins)

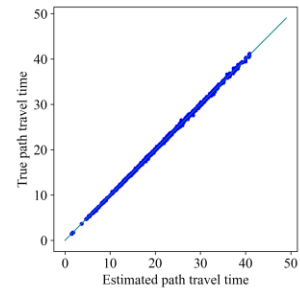

Fig. 16 The comparison between the estimated and the true OD path travel time (unit: mins)

(3) Deleting 20% OD pairs from the AFC data

On the basis of adding 20% white noise to the path travel time, we further randomly delete the records of 20% OD pairs. For the deleted OD pairs, the corresponding true average OD path travel time in the vector $t$ in Eqs. (4) is zero. Therefore, we use the "softimpute" method to conduct matrix completion. The estimated results are shown in Fig. 17 and Fig. 18. Both of the errors of the link travel time and station waiting time continue to increase. The estimated link travel time is still aligned with their corresponding true values. Results show that the proposed methodology is effective in this case.

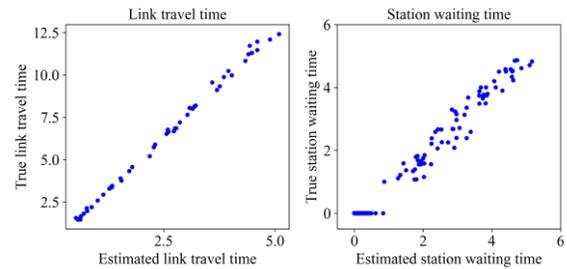

Fig. 17 The comparison between the estimated and the true time (unit: mins)

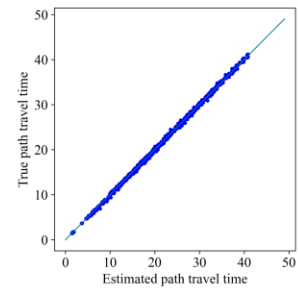

Fig. 18 The comparison between the estimated and the true OD path travel time (unit: mins)

(4) Deleting 50% OD pairs from the AFC data

On the basis of adding 20% white noise to the path travel time, we further randomly delete the records of 50% OD pairs. For the deleted OD pairs, the corresponding true average OD path travel time in the vector $t$ in Eqs. (4) is zero. Therefore, we use the "SoftImpute" method to conduct matrix completion. The estimated results are shown in Fig. 19 and Fig. 20. Both of the errors of the link travel time and station waiting time continue to increase. However, the errors of the station waiting time are too large to be acceptable. The errors of the estimated link travel time become larger in this case. Results show that the proposed methodology is not effective in this case.

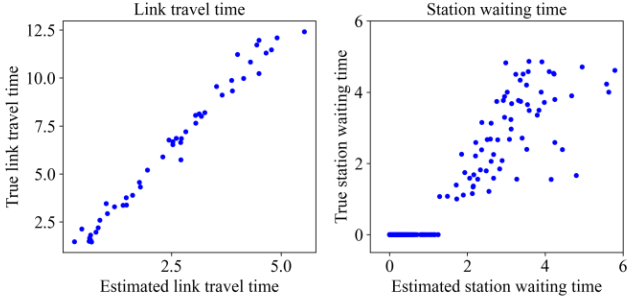

Fig. 19 The comparison between the estimated and the true time (unit: mins)

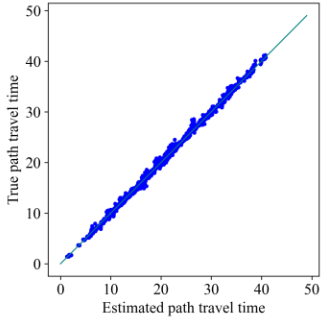

Fig. 20 The comparison between the estimated and the true OD path travel time (unit: mins)

## D. Result analysis for real-world subway networks

In this section, we will apply the methodology to the Beijing Subway network using real-world AFC data. For the real-world network, the true link travel time and the true station waiting time are unavailable. However, the true path travel time is available. The estimated $t$ includes the estimated link travel time and station waiting time in different time intervals. Leveraging these estimated link travel time and station waiting time, we can reconstruct the estimated path travel time. Therefore, the estimated path travel time is also available. The summary of this information is shown in Table 9.

Table 9 The information summary for the real-world subway network

|  | Estimated values | True values |
| --- | --- | --- |
| Link travel time | √ | × |
| Station waiting time | √ | × |
| Path travel time | √ | √ |

### 1) Convergence and goodness-of-fitting

In the real-world subway network, the comparison between the true and the estimated path travel time is shown in Fig. 21. The estimated values and the true values are generally aligned with each other. The $R^2$ reaches up to 0.700, which can show the effectiveness of the proposed methodology for the application in the real world.

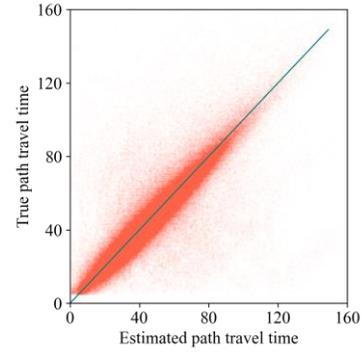

Fig. 21 The comparison between the estimated and the true OD path travel time (unit: mins)

### 2) Link travel time

The summary of the estimation results for the link travel time is shown in Fig. 22. There are nearly 250 links that cost small travel time (less than 1 minute), such as the transfer links in the same platforms. Most link travel time ranges from 1 minute to 5 minutes, which conforms to the real-world operation environments.

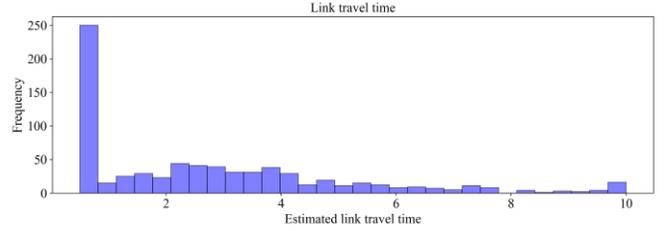

Fig. 22 The summary of the estimation results for the link travel time (unit: mins)

### 3) Station waiting time in different time intervals

The summary of the estimation results for the station waiting time is shown in Fig. 23. From the first time interval (07:00-07:30) to the tenth time interval (11:30-12:00), the number of the larger waiting time increases, while the number of the smaller waiting time decreases, which is in line with common sense. In the sixth time interval (09:00-09:30), the station waiting time seems to be the largest. We also select several specific stations to present their waiting time variations in different time intervals as shown in Fig. 24. The estimation results are also reasonable, which shows the effectiveness of the proposed methodology.

## V. CONCLUSION

In this study, we propose a CG-based framework to estimate the link travel time and station waiting time using the AFC data for large-scale URT networks. We first formulate a data-driven optimization model to estimate the link travel time and station waiting time. The optimization model is cast into a CG framework to obtain the estimation results efficiently. The proposed framework is strictly verified on a synthetic URT network and applied to a real-world URT network. To the best of our knowledge, this is the first time that the CG model is applied to the URT system. Several critical findings are summarized as follows.

(1) The CG framework is effective and efficient in the estimation of the link travel time and station waiting time

- under a URT network level.
(2) Sensitivity analysis indicates that the proposed CG-based methodology is robust to the data missing.
(3) The CG methodology can be verified using the AFC data. Therefore, the estimation results can be applied to real-world applications.

Overall, these findings can provide critical insights for the URT operation. For example, we can use the estimated travel time information reconstruct the real-time spatiotemporal distributions of passengers, which is essential for URT operation and management. However, there are also some limitations in this study. For example, we assume that the link travel time is not time dependent, which may not be true in some of the URT systems. Moreover, whether the CG-based methodology is effective for weekends is to be further explored.

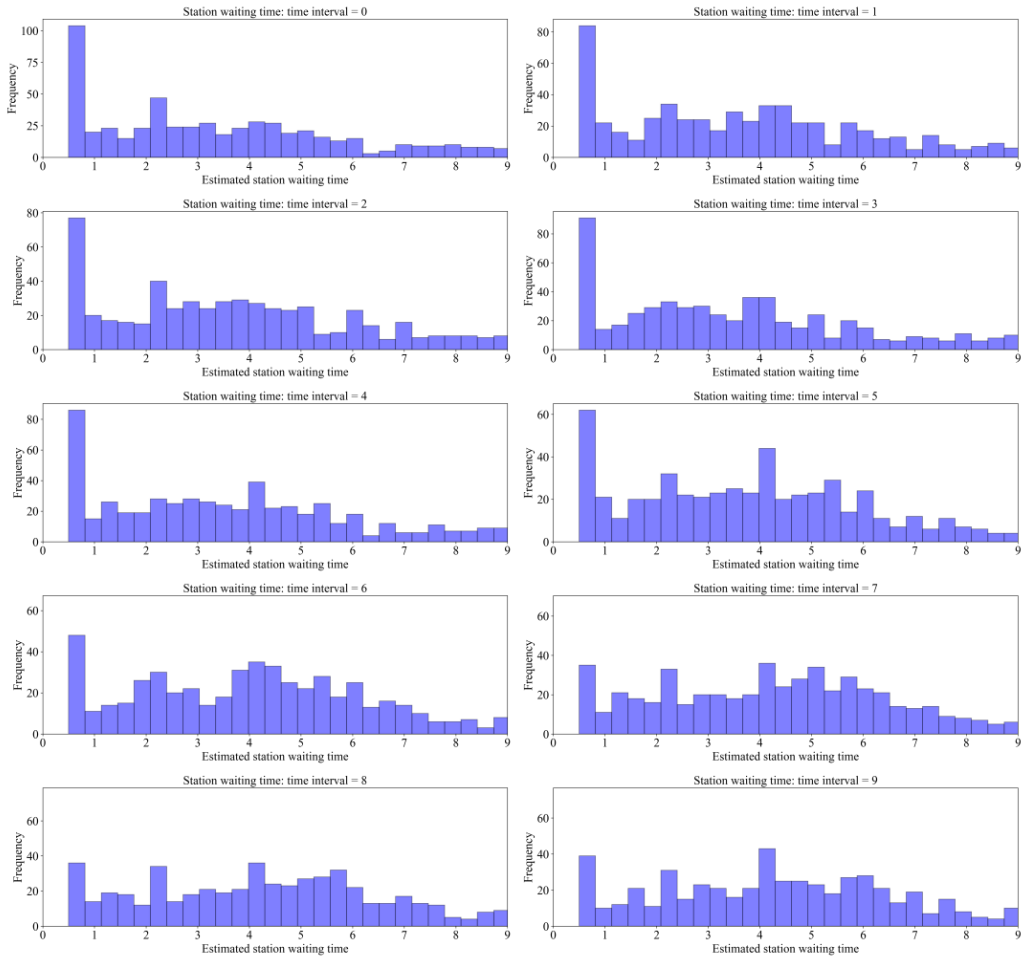

Fig. 23 The summary of the estimation results for the station waiting time in different time intervals

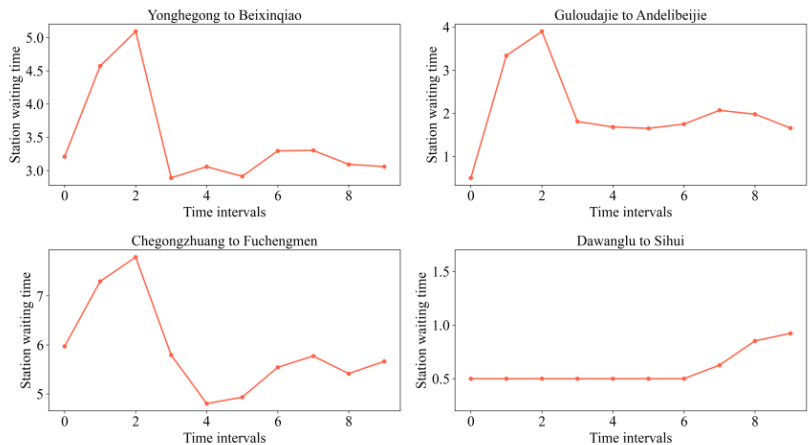

Fig. 24 The specific station waiting time variation in different time intervals

## VI. REFERENCES

[1] J. Zhang, H. Che, F. Chen, W. Ma, and Z. He, "Short-term origin-destination demand prediction in urban rail transit systems: A channel-wise attentive split-convolutional neural network method," *Transportation Research Part C: Emerging Technologies,* vol. 124, pp. 102928, 2021.

[2] J. Zhang, F. Chen and Q. Shen, "Cluster-based LSTM network for short-term passenger flow forecasting in urban rail transit," *IEEE Access,* vol. 7, pp. 147653-147671, 2019.

[3] Y. Zhu, H. N. Koutsopoulos and N. H. Wilson, "Passenger itinerary inference model for congested urban rail networks," *Transportation Research Part C: Emerging Technologies,* vol. 123, pp. 102896, 2021.

[4] Z. Dai, X. Ma and X. Chen, "Bus travel time modelling using GPS probe and smart card data: A probabilistic approach considering link travel time and station dwell time," *Journal of Intelligent Transportation Systems,* vol. 23, pp. 175-190, 2019.

[5] L. Yang, Y. Zhang, S. Li, and Y. Gao, "A two-stage stochastic optimization model for the transfer activity choice in metro networks," *Transportation Research Part B: Methodological,* vol. 83, pp. 271-297, 2016.

[6] J. Zhang, F. Chen, Y. Guo, and X. Li, "Multi-graph convolutional network for short-term passenger flow forecasting in urban rail transit," *IET Intelligent Transport Systems,* vol. 14, pp. 1210-1217, 2020.

[7] H. Lee, D. Zhang, T. He, and S. H. Son, "Metrotime: Travel time decomposition under stochastic time table for metro networks," in *2017 IEEE International Conference on Smart Computing (SMARTCOMP)*, 2017, pp. 1-8.

[8] F. Zhang, J. Zhao, C. Tian, C. Xu, X. Liu, and L. Rao, "Spatiotemporal segmentation of metro trips using smart card data," *IEEE Transactions on Vehicular Technology,* vol. 65, pp. 1137-1149, 2015.

[9] W. Li, X. Yan, X. Li, and J. Yang, "Estimate Passengers' Walking and Waiting Time in Metro Station Using Smart Card Data (SCD)," *IEEE Access,* vol. 8, pp. 11074-11083, 2020.

[10] B. Mo, Z. Ma, H. N. Koutsopoulosc, and J. Zhao, "Assignment-based Path Choice Estimation for Metro Systems Using Smart Card Data," *arXiv preprint arXiv:2001.03196,* 2020.

[11] C. Yu, H. Li, X. Xu, and J. Liu, "Data-driven approach for solving the route choice problem with traveling backward behavior in congested metro systems," *Transportation Research Part E: Logistics and Transportation Review,* vol. 142, pp. 102037, 2020.

[12] Y. Zhang, E. Yao, K. Zheng, and H. Xu, "Metro passenger's path choice model estimation with travel time correlations derived from smart card data," *Transportation Planning and Technology,* vol. 43, pp. 141-157, 2020.

[13] Y. Zhang, E. Yao, J. Zhang, and K. Zheng, "Estimating metro passengers' path choices by combining self-reported revealed preference and smart card data," *Transportation Research Part C: Emerging Technologies,* vol. 92, pp. 76-89, 2018.

[14] J. Zhao, F. Zhang, L. Tu, C. Xu, D. Shen, C. Tian, X. Li, and Z. Li, "Estimation of passenger route choice pattern using smart card data for complex metro systems," *IEEE Transactions on Intelligent Transportation Systems,* vol. 18, pp. 790-801, 2016.

[15] J. Wu, Y. Qu, H. Sun, H. Yin, X. Yan, and J. Zhao, "Data-driven model for passenger route choice in urban metro network," *Physica A: Statistical Mechanics and its Applications,* vol. 524, pp. 787-798, 2019.

[16] Y. Oh, Y. Byon, J. Y. Song, H. Kwak, and S. Kang, "Dwell Time Estimation Using Real-Time Train Operation and Smart Card-Based Passenger Data: A Case Study in Seoul, South Korea," *Applied Sciences,* vol. 10, pp. 476, 2020.

[17] L. Sun, Y. Lu, J. G. Jin, D. Lee, and K. W. Axhausen, "An integrated Bayesian approach for passenger flow assignment in metro networks," *Transportation Research Part C: Emerging Technologies,* vol. 52, pp. 116-131, 2015.

[18] F. Zhou and R. Xu, "Model of passenger flow assignment for urban rail transit based on entry and exit time constraints," *Transportation Research Record,* vol. 2284, pp. 57-61, 2012.

[19] X. Ma, C. Liu, H. Wen, Y. Wang, and Y. Wu, "Understanding commuting patterns using transit smart card data," *Journal of Transport Geography,* vol. 58, pp. 135-145, 2017.

[20] X. Ma, Y. Wu, Y. Wang, F. Chen, and J. Liu, "Mining smart card data for transit riders' travel patterns," *Transportation Research Part C: Emerging Technologies,* vol. 36, pp. 1-12, 2013.

[21] J. Zhao, Q. Qu, F. Zhang, C. Xu, and S. Liu, "Spatio-temporal analysis of passenger travel patterns in massive smart card data," *IEEE Transactions on Intelligent Transportation Systems,* vol. 18, pp. 3135-3146, 2017.

[22] J. Zhang, F. Chen, Z. Cui, Y. Guo, and Y. Zhu, "Deep Learning Architecture for Short-Term Passenger Flow Forecasting in Urban Rail Transit," *IEEE Transactions on Intelligent Transportation Systems,* pp. 1-11, 2020.

[23] J. Zhang, F. Chen, Z. Wang, and H. Liu, "Short-Term Origin-Destination Forecasting in Urban Rail Transit Based on Attraction Degree," *IEEE Access,* vol. 7, pp. 133452-133462, 2019.

[24] D. Yang, K. Chen, M. Yang, and X. Zhao, "Urban rail transit passenger flow forecast based on LSTM with enhanced long-term features," *IET Intelligent Transport Systems,* vol. 13, pp. 1475-1482, 2019.

[25] Z. Guo, X. Zhao, Y. Chen, W. Wu, and J. Yang, "Short-term passenger flow forecast of urban rail transit based on GPR and KRR," *IET Intelligent Transport Systems,* vol. 13, pp. 1374-1382, 2019.

[26] T. Yang, P. Zhao and X. Yao, "A Method to Estimate URT Passenger Spatial-Temporal Trajectory with Smart Card Data and Train Schedules," *Sustainability,* vol. 12, pp. 2574, 2020.

[27] L. Sun, D. Lee, A. Erath, and X. Huang, "Using smart card data to extract passenger's spatio-temporal density and train's trajectory of MRT system," in *Proceedings of the ACM SIGKDD international workshop on urban computing*, 2012, pp. 142-148.

[28] T. Kim, X. Zhou and R. M. Pendyala, "Computational Graph-based Framework for Integrating Econometric Models and Machine Learning Algorithms in Emerging Data-Driven Analytical Environments," *Transportmetrica A: Transport Science,* pp. 1-35, 2021.

[29] X. Wu, J. Guo, K. Xian, and X. Zhou, "Hierarchical travel demand estimation using multiple data sources: A forward and backward propagation algorithmic framework on a layered computational graph," *Transportation Research Part C: Emerging Technologies,* vol. 96, pp. 321-346, 2018.

[30] J. Sun, J. Guo, X. Wu, Q. Zhu, D. Wu, K. Xian, and X. Zhou, "Analyzing the Impact of Traffic Congestion Mitigation: From an Explainable Neural Network Learning Framework to Marginal Effect Analyses," *Sensors,* vol. 19, pp. 2254, 2019-05-15 2019.

[31] W. Ma, X. Pi and S. Qian, "Estimating multi-class dynamic origin-destination demand through a forward-backward algorithm on computational graphs," *Transportation Research Part C: Emerging Technologies,* vol. 119, pp. 102747, 2020.

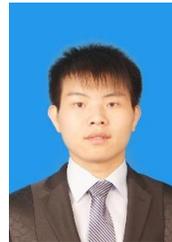

**Jinlei Zhang** was born in Hebei Province, China. He received the Ph.D. degree from Beijing Jiaotong University, China. He is currently an assistant professor at Beijing Jiaotong University.

His research interests include machine learning, deep learning, traffic data mining and applications, and dynamic traffic modeling and management.

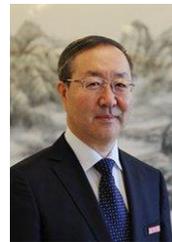

**Feng Chen** received his B.S. degree, M.A. degree, and Ph.D. degree in railway engineering from Beijing Jiaotong University, Beijing, China, in 1983, 1990, and 2007, respectively. He is currently working as a professor at Beijing Jiaotong University and is the president of China University of Petroleum.

His research interests include passenger flow management, traffic data mining, and application for urban rail transit. Professor Chen won the first prize of national science and technology progress as the second accomplisher in 2017.

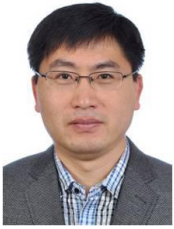
**Lixing Yang** received the B.S. and M.S. degrees from the Department of Mathematics, Hebei University, Baoding, China, in 1999 and 2002, respectively, and the Ph.D. degree from the Department of Mathematical Sciences, Tsinghua University, Beijing, China, in 2005.

Since 2005, he has been with the State Key Laboratory of Rail Traffic Control and Safety, Beijing Jiaotong University, Beijing, where he is currently a Professor. He is the author or coauthor of more than 80 papers published in national conferences, international conferences, and premier journals. His current research interests include stochastic programming, fuzzy programming, intelligent systems, and applications in transportation problems and rail traffic control systems.

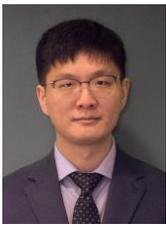
**Wei Ma** received bachelor's degrees in Civil Engineering and Mathematics from Tsinghua University, China, master degrees in Machine Learning and Civil and Environmental Engineering, and PhD degree in Civil and Environmental Engineering from Carnegie Mellon University, USA. He is currently an assistant professor with the Department of Civil and Environmental Engineering at the Hong Kong Polytechnic University (PolyU).

His research focuses on the intersection of machine learning, data mining, and transportation network modeling, with applications for smart and sustainable mobility systems. He has received awards for research excellence and his contributions to the area, including 2020 Mao Yisheng Outstanding Dissertation Award, and best paper award (theoretical track) at INFORMS Data Mining and Decision Analytics Workshop.

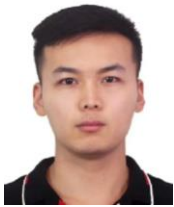
**Guangyin Jin** is a Ph.D. candidate at College of Systems Engineering of National University of Defense Technology.

His research interest falls in the area of spatial-temporal data mining, urban computing and intelligent transportation.

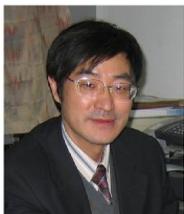
**Ziyou Gao** received the Ph.D. degree from the Institute of Applied Mathematics, Chinese Academy of Sciences, Beijing, China, in 1994.

He is currently a Professor with the State Key Laboratory of Rail Traffic Control and Safety, Beijing Jiaotong University. He also serves as the President of the Society of Management Science and Engineering of China, the Vice-President of the Systems Engineering Society of China, and the Vice-President of the Chinese Society of Optimization, Overall Planning and Economic Mathematics. In 2003, he was elected as a foreign member of the Russian Academy of Natural Sciences. His research interests include urban traffic management, management and optimization of urban rail transit, complexity of transportation network, and transport modeling and planning.